# MS and PAN image fusion by combining Brovey and wavelet methods

**Hamid Reza Shahdoosti**

Department of Electrical Engineering, Hamedan University of Technology

**Abstract**
Among the existing fusion algorithms, the wavelet fusion method is the most frequently discussed one in recent publications because the wavelet approach preserves the spectral characteristics of the multispectral image better than other methods. The Brovey is also a popular fusion method used for its ability in preserving the spatial information of the PAN image. This study presents a new fusion approach that integrates the advantages of both the Brovey (which preserves a high degree of spatial information) and the wavelet (which preserves a high degree of spectral information) techniques to reduce the colour distortion of fusion results. Visual and statistical analyzes show that the proposed algorithm clearly improves the merging quality in terms of: correlation coefficient and UIQI; compared to fusion methods including, IHS, Brovey, PCA , HPF, discrete wavelet transform (DWT), and a-trous wavelet.

## Introduction

The satellites observing the earth provide two types of images: several high spectral/low spatial resolution multispectral (MS) images and a low-spectral/high spatial panchromatic (PAN) image. This complementary information are used by image fusion methods (or pan-sharpening methods) to obtain a fused image, having a high spectral/high spatial resolution, which is more informative and interpretative than both initial images.

The most commonly used image-fusion methods are those based on the IHS (also known as HSI) [1], [2] and principal component analysis (PCA) [3], [4]. However, these methods can cause spectral distortion in the results [2]. Chavez proposed the HPF(High Pass Filtering) fusion algorithm which has shown better performance in terms of the high-quality synthesis of spectral information [5], [6]. The principle of this method is based on extracting the high-frequency information from the PAN image and injecting it into the MS image. The Box car filter is used for extracting the high-frequency information in this method. However, the ripple in the frequency response of box car filters has some negative impact on the fusion results.

The well-known Mallat's fusion algorithm [7] uses an orthonormal basis and it can effectively preserve spectral information, but because of down-sampling operators which are used to implement ordinary discrete wavelet transforms, these transforms are not shift invariant and they can lead to problem in data fusion [8]. To avoid the above-mentioned problem, the discrete wavelet transform known as "à trous" algorithm [9-11] was proposed by eliminating the decimation operators in the wavelet structure. It is a shift-invariant and redundant wavelet transform algorithm based on a multiresolution dyadic scheme. Similar to Mallat's fusion algorithm, à trous can preserve the spectral information of MS images well.

The Brovey method is based on the chromaticity transform [13], [14], [15]. It is a simple but powerful method for combining the data acquired by different sensors. The Brovey preserves the spatial information of PAN image well. However distortion of the spectral information is not acceptable in this method.

To overcome the deficiency of the Brovey method, a new fusion approach is proposed in this paper. This new method integrates the advantages of both the Brovey (which preserves a high degree of spatial information) and the à trous (which preserves a high degree of spectral information) techniques to preserve both spatial and spectral information of PAN and MS images. To verify the efficiency of the proposed method, visual and statistical assessments are carried out on the MS and PAN data.

## À trous algorithm

This section reviews the à trous fusion method as presented in [11],[15],[16] and briefly discusses the characteristics of these method. Two different strategies can be used in the à trous fusion method which are the additive wavelet (AW) and substitution wavelet (SW). The AW injects the high frequency wavelet planes of the PAN image directly into the MS image. Thus, the radiometric signature of the MS image is changed by this algorithm. This means that the AW method produces redundant geometric information. This method can be formulated as follow:

$$F_i = MS_i + \sum_{j=1}^{n} w_j \qquad (1)$$

where $w_j$ (j=1,…,n) is the high frequency wavelet plane of the PAN image, $F_i$ is the i[th] fused band and $MS_i$ is the i[th] band of the multispectral image.

The SW substitutes the high frequency wavelet planes of the MS image for the high frequency wavelet planes of the PAN image as follow:

$$F_i = MS_i + \sum_{j=1}^{n} w_j - \sum_{j=1}^{n} v_{j,i} \quad (2)$$

where $v_{j,i}$ (j=1,…,n) is the high frequency wavelet plane of the i$^{th}$ band of the MS image.

The wavelet planes of the MS image are discarded in this method. This implies that, the local high frequency spatial information that is visible in the MS image can be missing in the fused image. Therefore, the SW method can eliminate both the geometric and radiometric information of the MS image. So the injection of high frequency wavelet planes in the à trous algorithm is a challenging problem.

**Brovey algorithm**

The Brovey fuses the PAN and MS images as following [17-19]:

$$F_i = \frac{PAN}{M} MS_i \quad (3)$$

where M is a linear combination of MS images in the Brovey method. So we will have:

$$F_i = \frac{PAN}{\frac{1}{N}\sum_{i=1}^{N} MS_i} MS_i \quad (4)$$

The spatial information is well preserved in this method but this method leads to the spectral distortion in the results [17]. The Brovey method takes into account the relative radiometric signature of the MS bands and preserves local high frequency spatial information of MS image.

**Proposed method**

In the past few years, several researchers proposed different PAN and MS image-fusion methods based on the hybrid concept to integrate the advantage of two fusion method. Undecimated Wavelet PCA (dWPC) [15], undecimated Wavelet IHS (dWI) [15], IHS-wavelet [2], IHS-retina-inspired [20], and FFT-enhanced IHS [21] are the most commonly used hybrid methods. As discussed in the previous sections, the Brovey algorithm can preserve high degree of spatial information of PAN image and *à trous* can effectively maintain spectral information of MS image. We propose Brovey-Wavelet (BW) method to obtain a fused image with simultaneously high spectral and spatial resolution. The steps of the proposed BW method are the following.

1- Resample the MS image to make its pixel size equal to that of the PAN image.
2- Compute the $F_i$ by the Brovey algorithm (Equation (4)).
3- Decompose $F_i$ and $MS_i$ to n wavelet planes (n is the resolution ratio between PAN and MS)
4- Substitute the high frequency wavelet plane of the $MS_i$ images by the high frequency wavelet planes of $F_i$ (without any histogram matching).
5- Perform the inverse wavelet transform.

The proposed method injects the spatial information, while considering the relative radiometric signature of the MS bands. In addition, it preserves local high frequency parts of the spatial information of MS images. This hybrid method can significantly improve the fusion quality in terms of both visual assessment and quantitative assessment.

## Experimental results

Four-band multispectral QuickBird data was used for our experiments. The raw images are downloaded from http://studio.gge.unb.ca/UNB/images. These images are acquired by a commercial satellite, QuickBird. This data set was taken over the Pyramid area of Egypt in 2002. The subscenes of the raw images are used as PAN and MS images.

First we assess the proposed method visualy. The visual performances of the fused images are shown in Figure 1. As can be seen from Figure 1, the spatial details pretty good enhanced in the merged image using SW method, but the colors have changed clearly. Another disadvantage is that small objects miss their spectral content (red window in Figure 1, shows one of the objects as an example).

As can be seen from Figure 1, the sharpening in the AW method is not suitable, however it is difficult to discuss about spectral information in this figure.

Figure 2 shows fusion results of SW, AW and the proposed method for the data set obtained by the Ikonos satellite. In this figure we can see over edge enhancement of SW method which is reported frequently [17],[22]. Spectral distortion of SW method can be seen more easier from this image. In AW method, sharpening of most areas such as green trees is not enough.

As can be seen from Figures 1 and 2, the BW method does not have over sharpening problem. It preserves the color of small objects. In addition, the colors of fused image are closer to the multispectral images in our method. So we have a fused image with simultaneous high spectral and spatial resolution by the BW merger.

In addition to the visual inspection, the performance of each method is analyzed quantitatively. In order to assess the quality of the merged images at the inferior level, two objective indicators were used.

1) CC

This metric indicates the degree of linear dependence between the original reference and fused images [11]. If two images are identical, the correlation coefficient will be maximal and equals 1. It is defined as follows:

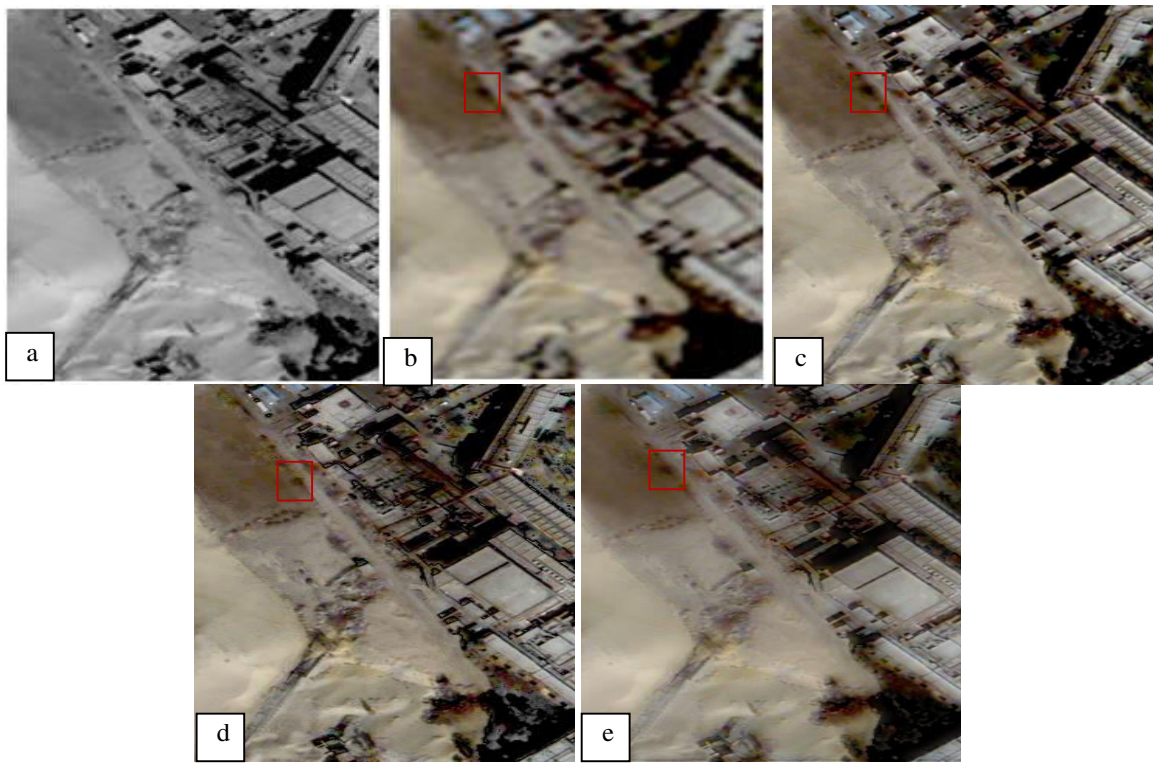

Figure(1)- (a) PAN. (b) MS. (c) Fused image by the proposed method. (d) Fused image by SW. (e) Fused image by AW.

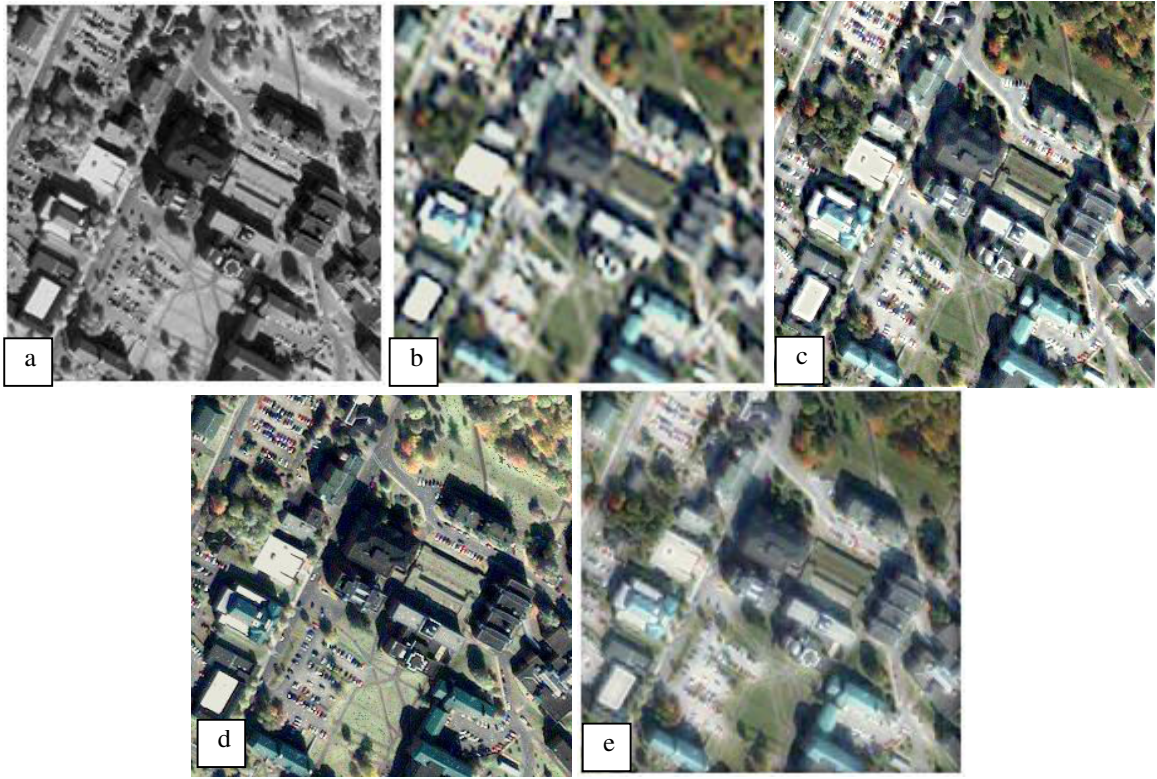

Figure (2)- (a) PAN. (b) MS. (c) Fused image by the proposed method. (d) Fused image by SW. (e) Fused image by AW.

Table 1. CC computed between fused MS images and reference MS images using 8*8 window

| CC | IHS | Brovey | PCA | HPF | AW | SW | dWI | dWPC | BW |
|---|---|---|---|---|---|---|---|---|---|
| Red | 0.7881 | 0.7656 | 0.7972 | 0.8008 | 0.8210 | 0.8358 | 0.8942 | 0.8860 | 0.9110 |
| Green | 0.7673 | 0.7510 | 0.7760 | 0.8114 | 0.8132 | 0.8540 | 0.8527 | 0.8644 | 0.8988 |
| Blue | 0.7943 | 0.8078 | 0.7993 | 0.8217 | 0.7821 | 0.7930 | 0.8026 | 0.8030 | 0.8429 |
| NIR | 0.7423 | 0.8133 | 0.8046 | 0.8346 | 0.8530 | 0.8103 | 0.8715 | 0.8874 | 0.9001 |
| Mean | 0.7730 | 0.7844 | 0.7943 | 0.8171 | 0.8173 | 0.8233 | 0.8553 | 0.8602 | 0.8882 |

Table 2. UIQI computed between fused MS images and reference MS images using 8*8 window

| UIQI | IHS | Brovey | PCA | HPF | AW | SW | dWI | dWPC | BW |
|---|---|---|---|---|---|---|---|---|---|

| Red   | 0.7695 | 0.7445 | 0.7825 | 0.7860 | 0.8112 | 0.8139 | 0.8706 | 0.8760 | 0.9023 |
| Green | 0.7309 | 0.7243 | 0.7614 | 0.7904 | 0.8067 | 0.8389 | 0.8437 | 0.8478 | 0.8771 |
| Blue  | 0.7505 | 0.7719 | 0.7753 | 0.8012 | 0.7469 | 0.7716 | 0.7775 | 0.7782 | 0.8074 |
| NIR   | 0.7008 | 0.7848 | 0.7816 | 0.8088 | 0.8426 | 0.8014 | 0.8420 | 0.8502 | 0.8674 |
| Mean  | 0.7379 | 0.7564 | 0.7752 | 0.7966 | 0.8019 | 0.8065 | 0.8335 | 0.8380 | 0.8635 |

$$\mathrm{Corr}(A,B) = \frac{\sum_{j=1}^{npix}(A_j - m_A)(B_j - m_B)}{\sqrt{\sum_{j=1}^{npix}(A_j - m_A)\sum_{j=1}^{npix}(B_j - m_B)}} \quad (5)$$

2) UIQI

It is defined as follows [23]:

$$UIQI(A,B) = \frac{\sigma_{AB}}{\sigma_A \sigma_B} \cdot \frac{2 m_A m_B}{m_A^2 + m_B^2} \cdot \frac{2 \sigma_A \sigma_B}{\sigma_A^2 + \sigma_B^2} \quad (6)$$

where *A* and *B* are fused and reference images respectively. The universal image quality index (UIQI) models any distortion as a combination of three different factors: loss of linear correlation, contrast distortion and luminance distortion.

As we know, signals which are obtained by the image sensors are not stationary in general so image quality is often space variant, but in practice it is desired to evaluate an entire image using a single overall quality value. Therefore, it is more rational to measure indexes locally and then combine them together. We apply CC and UIQU quality metrics to local regions using a sliding window approach. Starting from the top-left corner of the image, a sliding window of size 8*8 moves horizontally and vertically pixel by pixel through all the rows and columns of the image until the bottom-right corner is reached. Thereafter, we compute the average of these quality indexes to obtain the overall quality value.

The Tables 1 and 2 show the values of CC and UIQI respectively for the proposed method and other popular methods.

As can be seen from Tables 1 and 2, the quality indexes obtained by applying the proposed method are pretty good compared with those obtained by applying other methods. These statistical assessment are in harmony with the visual analysis.

**Conclusions**

The Brovey method considers the relative radiometric signature of the MS bands and preserves local high frequency spatial information of MS image. The SW method maintains low frequency spectral information of MS image. In this paper, the BW method was proposed to take the advantages of both methods. Finally, the visual results show that BW can achieve better performance relative to SW and AW methods. In addition to the visual inspection, the performance of proposed method and some well-known methods was analyzed quantitatively. We applied CC, and UIQI metrics locally and computed the average of these local indexes to obtain the final quality values. The quality values show that the BW method can achieve a better performance.